\documentclass[letterpaper,twocolumn,10pt]{article}
\usepackage{usenix,epsfig,endnotes}

\usepackage[utf8]{inputenc}
\usepackage{color}
\usepackage{cite}
\usepackage{url}
\usepackage{graphicx}
\usepackage{authblk}
\usepackage{xspace}
\usepackage{array}
\usepackage{mathtools}
\usepackage{amsmath, amsthm, amssymb}
\usepackage{subcaption}
\usepackage{booktabs}
\usepackage{fancyhdr}
\usepackage{theoremref}

\newtheorem{theorem}{Theorem}
\newtheorem*{theorem*}{Theorem} 

\newcommand{\etal}[0]{\emph{et al.}\xspace}
\newcommand{\eg}[0]{\emph{e.g.,}\xspace}

\newif\ifsubmit
\submitfalse

\ifsubmit
\newcommand{\kevin}[1]{}
\else
\newcommand{\kevin}[1]{\textcolor{cyan}{Kevin: #1}}
\fi

\newcommand{\ignore}[1]{}

\begin{document}

\date{}

\renewcommand{\headrulewidth}{0pt}

\title{\Large \bf Designing Adversarially Resilient Classifiers using Resilient Feature Engineering}

\author{
{\rm Kevin Eykholt, Atul Prakash}\\
University of Michigan\\
} 

\maketitle

\begin{abstract}
We provide a methodology, resilient feature engineering, for creating adversarially resilient classifiers. According to existing work, adversarial attacks identify weakly correlated or non-predictive features learned by the classifier during training and design the adversarial noise to utilize these features. Therefore, highly predictive features should be used first during classification in order to determine the set of possible output labels. Our methodology focuses the problem of designing resilient classifiers into a problem of designing resilient feature extractors for these highly predictive features. We provide two theorems, which support our methodology. The Serial Composition Resilience and Parallel Composition Resilience theorems show that the output of adversarially resilient feature extractors can be combined to create an equally resilient classifier. Based on our theoretical results, we outline the design of an adversarially resilient classifier.
\end{abstract}

\section{Introduction}
Deep neural networks and other machine learning algorithms have seen large success being used in industry for numerous tasks. However, the rise in interest towards these algorithms has also raised concerns in the community as to their trustworthiness. Namely, these algorithms are vulnerable to \textit{adversarial inputs}. These are inputs that, to a human, appear no different from an existing input the algorithm can correctly label, but are slightly modified such that the algorithm assigns a different label to them. These types of inputs can harm the widespread adoption of machine learning algorithms, especially in safety critical tasks. For example, in autonomous driving, machine learning is used to perform object recognition, thus it is critical that the algorithm is robust to such adversarial inputs.

Previous research in adversarial machine learning limited attacks to the digital domain. Adversarial modifications would be directly applied to the input of the algorithm (\eg an image file). Recent work, however, has demonstrated that it is possible to craft adversarial inputs in the physical domain~\cite{rp2, DBLP:journals/corr/abs-1804-05810}. Thus, it is critically important to design defensive measures that can reduce or remove the risk of adversarial examples. In the digital domain, many defensive solutions have been proposed, but most strategies are limited. As pointed out by previous work, most defensive solutions only work on a static adversary or under very specific conditions \cite{carlini2017towards,DBLP:journals/corr/abs-1804-03286,DBLP:journals/corr/CarliniW17, DBLP:journals/corr/abs-1802-00420}.

Current literature identifies that adversarial examples are possible due to machine learning models relying on features which are weakly correlated with the correct output label~\cite{2018robustness} or assigning non-zero weights to irrelevant features~\cite{2018explainability}. Thus, classifiers should first use the highly predictive features to narrow down the set of possible output labels. These features are likely to be highly selective and in certain domains, may be easily identifiable. For example, in road sign classification, knowing the shape of the sign can be enough to uniquely identify the sign or sign class (\eg Octagon: Stop sign; Diamond: Warning sign).

In our work, we propose \textit{resilient feature engineering}, which uses highly predictive features to create adversarially resilient machine learning classifiers. First, predictive features for the classification task are identified. For each identified feature, a feature extractor is constructed. The output of each feature extractor is provided to a classification algorithm, which outputs one or more possible class labels based on the task.

The key idea is to create these feature extractors so they are \textbf{resilient} to adversarial attacks. That is to say, if only a certain amount of input distortion is allowed, the feature extractors cannot be adversarially attacked.  We provide two theorems, the Serial Composition Resilience and Parallel Composition Resilience theorems, that state if such conditions can be maintained, then the classifier is also adversarially resilient. We discuss a basic design for an adversarially resilient classifier based on our theoretical results.

\section{Related Works}

The existence of adversarial examples was first studied by Szegedy \etal \cite{szegedy2014intriguing}. In their work, they found that it is possible to arbitrarily manipulate the output of a neural network using imperceptible, non-random perturbations to the input, denoting these perturbed inputs as \textit{adversarial examples}. Since then, numerous works have appeared proposing new algorithms for generating adversarial examples in both the digital domain \cite{papernot2016limitations,goodfellow2014explaining,carlini2017towards,moosavi2015deepfool} and the physical domain \cite{rp2, DBLP:journals/corr/abs-1804-05810, kurakin2016adversarial, athalye2017synthesizing, rp2objectdetector}.

In an effort to mitigate the effect of adversarial examples, there is a push to design defensive measures against adversarial examples. Research in this area can be divided into defensive methods \cite{goodfellow2014explaining, distillation, tramer2018ensemble, song2018pixeldefend}, which improve a model's resistance to adversarial examples, and detection solutions \cite{DBLP:journals/corr/HendrycksG16b,DBLP:journals/corr/LiL16e, DBLP:journals/corr/GongWK17}, which use statistical properties or other information to identify adversarial examples from natural examples. 

The main issue with many of the proposed strategies is that many of them do not provide provable security guarantees assuming a certain threat model. Rather, they implement a technique, such as gradient masking, PCA analysis, or input normalization, and show that in testing the proposed strategy is resistant to many adversarial attacks. However, when exposed to an adaptive adversary that modifies the attack based on the defense, the defense fails \cite{carlini2017towards,DBLP:journals/corr/abs-1804-03286,DBLP:journals/corr/CarliniW17, DBLP:journals/corr/abs-1802-00420}. 

Of interest is research that provides provable security guarantees against adversarial attacks. Hein and Andriushchenko provided the first formal guarantees on the adversarial robustness of a classifier \cite{DBLP:journals/corr/HeinA17}. More specifically, given a specific instance, they demonstrate a lower-bound on the $l_{2}$ norm of the input manipulation required to change the output of the classifier. Similarly, Raghunathan \etal also establish lower bounds on adversarial perturbations, but do so for the $l_\infty$ norm \cite{raghunathan2018certified}.

Concurrently, Wong and Kolter also provide an adversarial robust training methodology for arbitrarily deep ReLU networks given a certain norm-bound on adversarial perturbations \cite{DBLP:journals/corr/abs-1711-00851}. In their work, they establish a convex space around each input and ensure that the classification decision for correctly labelled inputs does not change within the space. Then, optimize the convex space on the training point with the highest training loss. Finally, Sinha \etal also provide methods to guarantee adversarial robustness, but their method uses defined distributional Wasserstein distances rather than the norm of the adversarial perturbations \cite{sinha}.

Compared to previous work, our work provides a more general methodology for generating adversarially resilient classifiers. Our approach transforms the problem of defending against adversarial attacks into designing resilient feature extractors. Furthermore, we can leverage previous results from other works to inform the design of resilient feature extractors, rather than attempting to apply the results of those works to an entire classifier.

\section{Formal Background Definitions}
In this section, we formally define classifiers and adversarial inputs as well as some terms related to them. These definitions are necessary before discussing resilient feature engineering.

\subsection{Classifiers}
\label{sec:Classifiers}
Informally, a classifier is given a set of inputs and generates a set of outputs. Each output represents the class label for a given input where the class label belongs to a set of class labels. For example, a image classifier may be provided a picture of an animal and is expected to output the type of animal that is in the picture.

We define $x$, the input to a classifier, as a set of observations. An \textbf{observation} is a categorical, ordinal, integer valued, or real valued property. We define $y$, the output of a classifier, as a \textbf{label} or \textbf{category} belonging to a set of labels $Y$. Typically, $y$ is a single value, but it may be a tuple.

With these two preliminary definitions, we say that a \textbf{classifier} $F$ is a function that maps a set of observations $x$ to a label $y$ in the set of labels $Y$. That is to say:
 \begin{align*}
     F(x) = y
 \end{align*}
 
 In order to determine the correctness of a classifier, there must be some source that generates the correct label for a given input. Given a classifier $F$, $O_{F}$ denotes the \textbf{oracle} function for the classifier $F$. For an input $x$, $O_{F}(x) = l$ where $l \in Y \cup non$. $non$ is a label that is not contained in $Y$. If $O_{F}(x) = non$, then we say that x is a \textbf{nonsense input}. If $O_{F}(x) \neq non$, then $l \in Y$ and we say that x is a \textbf{natural input} and is part of the \textbf{natural domain} of $F$, denoted by $N_F$.
 
A nonsense input is an input in which no label in $Y$ would be appropriate. That is to say, the input is not suitable for the classification task. For example, if we treat a human labeller as the oracle, if given a picture of a dog and asked to label it as ``stop sign'' or ``cup'', the human labeller would  be unable to make a correct decision. None of the labels in the label set can be reasonably associated with the input.

For natural inputs, the output of the oracle for a given $x$ is referred to as the \textbf{true label} of $x$. 

With the formal definition of an oracle, we can now verify the correctness of a classifier. Given an input $x \in N_F$, $F(x)$ is \textbf{correct} if and only if $F(x) = O_F(x)$.  Otherwise, $F$ is \textbf{incorrect}. If $\forall x \in N_F, F(x) = O_F(X)$, then we say that $F$ is \textbf{perfectly accurate}. 

We will denote the space of inputs correctly labelled by $F$ as $C_F$ and the space of inputs incorrectly labelled by $F$ as $I_F$. Therefore, $N_F = C_F \cup I_F$.

\subsection{Adversarial Inputs}
Informally, an adversarial input to a classifier is an input that is maliciously designed to be mislabeled by the classifier, but would be obvious for a human to label correctly. Furthermore, adversarial inputs typically appear almost identical to a correctly labelled input.

For a classifier $F$, we say that an input $x \in I_F$ is \textbf{$\lambda$-adversarial} if $\exists \gamma: |\gamma| \leq \lambda$ and $F(x+\gamma) = O_F(x+\gamma)$ and  $O_F(x+\gamma) = O_F(x)$. Given a $\lambda$, we denote the space of $\lambda$-adversarial inputs for $F$ as $A^\lambda_F$.

In current literature, $\gamma$ is a measure of \textbf{distortion}, based on some measure of distance between two inputs. We use the addition operator (\eg $x+\gamma$), to represents changes made to an input $x$ by some amount of distortion $\gamma$. $\lambda$ is an upper bound on the distortion allowed when creating an adversarial input. It is typically small so adversarial inputs appear similar to correctly labelled inputs. For example, if the input is an image, the $\gamma$  would be a vector of positive and negative pixel value modifications.  A common choice for measuring $|\gamma|$ for image inputs is computing the $L_p$ norm of $\gamma$.

Given a classifier $F$, we that $F$ is \textbf{$\lambda$-resilient} if and only if $A^\lambda_F = \emptyset$. Furthermore, if for all $\lambda > 0, A^\lambda_F = \emptyset$, then we say that $F$ is \textbf{perfectly resilient}. That is to say, there are no adversarial examples for $F$.

\section{Resilient Feature Engineering}
In this section, we present the two main theorems, which formally support resilient feature engineering. After which, we introduce resilient feature engineering and discuss the implications and use of the theorems.

\ignore{
\begin{theorem}
\thlabel{thm:papr}
 Let $F$ be a classifier. If $F$ is perfectly accurate, then $F$ is perfectly resilient.
\end{theorem}
The proof for this theorem can be found in the appendix. If it is possible to design a classifier that is always correct for any input in its natural domain, then it is possible to design a classifier that cannot be adversarially manipulated.
}

\subsection{Composition Resilience Theorems}
We now present the Serial Composition Resilience and Parallel Composition Resilience theorems. The proof for both theorems can be found in the appendix.  The Serial Composition Resilience theorem is as follows:
\begin{theorem}
\thlabel{thm:2class}
If a classifier function $F(\cdot)$ is $\lambda$-resilient for some $\lambda$, then  classifier function defined as $G(F(\cdot))$ is also $\lambda$-resilient.
\end{theorem}

The Serial Composition Resilience theorem demonstrates that one can use the output of a resilient classifier function as the input to another classifier function resulting in an equally resilient classifier function. Note that although we use the term ``classifier'', our definition of classifier function allows use to extend these results to other functional constructs. A feature extractor, for example, is a classifier function by definition. It takes in an input and produces an output belonging to a set of feature labels.

The Parallel Composition Resilience theorem is as follows:
\begin{theorem}
\thlabel{thm:nclass}
 If there are multiple classifiers $F_1(\cdot), F_2(\cdot)...F_n(\cdot)$ which are all $\lambda$-resilient for some $\lambda$, then the composition $G(<F_1(\cdot), F_2(\cdot)...F_n(\cdot)>)$ is also $\lambda$-resilient.
\end{theorem}

The Parallel Composition Resilience theorem demonstrates that one can use the output of multiple resilient classifier functions as the input to another classifier function resulting in an equally resilient classifier function. This theorem extends the results of the Serial Composition Resilience theorem allowing us to join multiple resilient feature extractors as part of the classification decision. Often, using multiple feature can increase the selectivity  for classification. For example, suppose $F$ extracts the dominant sign color and $G$ maps the color to set of road sign containing the color. Given the color ``red'', $G$ could output \{Stop sign, Do Not Enter sign\}, which are both road signs that are predominantly red. However, adding a shape extractor would allow one to differentiate between a Stop sign and a Do Not Enter sign.

These two theorems allow us to simplify the task of creating an adversarially resilient classifier into building adversarially resilient feature extractors. The model designer would first identify predictive features for the classification task either using pre-existing domain knowledge or through some automatic discovery process. The difficulty remains in designing the feature extractors may be an easier task. If enough features are used, it may be possible to uniquely label every possible input.


\subsection{Basic Design}
Resilient feature engineering is composed of the following steps:
\begin{enumerate}
    \item Identify a feature for the classification task
    \item Create a $\lambda$-resilient feature extractor for the identified feature
    \item Repeat steps 1 and 2 until no new features are identified or the selectivity given by the features does not increase
    \item Provide the output of the feature extractors as an input to a classifier, which maps the features to the label set for the classification task
\end{enumerate}

\begin{figure}[t]
  \center
  \includegraphics[width=0.25\textwidth]{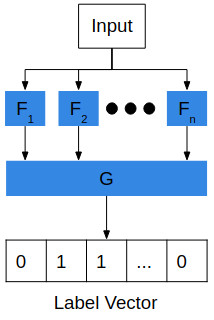}
  \caption{The basic implementation of a resilient classifier built using resilient feature engineering. The output is a vector of 0's or 1's with dimensions equal to the number of possible output classes. A 1 in index $i$ means that class label $i$ is a possible label for the input based on the extracted features.}
  \label{fig:basic}
\end{figure}

In this paper, we assume feature identification is based on pre-existing domain knowledge. Each identified feature should result in an increase in selectivity for the final classifier. Otherwise, the feature is redundant. The basic model architecture implied by the two theorems is shown in Figure \ref{fig:basic}. 

First, the input $x$ is processed by each of the  $\lambda$-resilient feature extractors. Then, the set of extracted features for that input is provided to a different classifier $G$, which outputs a vector containing one or more non-zero values. If the extracted features are sufficient to uniquely identify the label for $x$, then only one non-zero value is present. Otherwise, multiple non-zero values may be present indicate multiple possible output labels due to shared feature sets between labels. The weighting of the non-zero values in the final output vector is a design decision. In Figure \ref{fig:basic}, we assume equal weighting, but it could be based on some contextual information. Using unequal weighting may be desirable, especially if the output vector is intended to be used in a future computation.

We will use a basic traffic sign classification task to illustrate how the basic implementation works. Suppose, there are two resilient feature extractors, $F_1$ and $F_2$, that extract the dominant color and shape of a sign in the input respectively. $G$ can be any classification algorithm, but for simplicity we will assume it is a decision tree with the possible output vector values pre-determined based on existing domain knowledge. Furthermore, we will limit the possible output labels to the following list of U.S. stop signs\footnote{The designation and appearance of the signs can be found at \url{https://mutcd.fhwa.dot.gov/services/publications/fhwaop02084/}}: 

\begin{itemize}
    \item{Regulatory: Stop, Yield, Do Not Enter}
    \item{Warning: Left Turn Ahead, Right Turn Ahead}
    \item{Pedestrian: No Pedestrians}
    \item{Other: Speed Limit 25, Speed Limit 45, Hospital}
\end{itemize}

Therefore, the output vector has 9 indices with the labels ordered based on the above list. We denote the combined classifiers $F$ and $G$ as $R$ for convenience. Now, consider the following input signs: Stop, Left Turn Ahead, Hospital. The resilient classifiers $F_1$ and $F_2$ would have the following outputs:
\begin{itemize}
    \item{Stop: $F_1(x) =$ Red, $F_2(x) =$ Octagon}
    \item{Left Turn Ahead: $F_1(x) =$ Yellow, $F_2(x) =$ Diamond}
    \item{Hospital: $F_1(x) =$ Blue, $F_2(x) =$ Square}
\end{itemize}

and the output of $G$ based on the extracted features would be:
\begin{itemize}
    \item{Stop: $G(x) = <1,0,0,0,0,0,0,0,0>$}
    \item{Left Turn Ahead: $G(x) = <0,0,0,1,1,0,0,0,0>$}
    \item{Hospital: $G(x) = <0,0,0,0,0,0,0,0,1>$}
\end{itemize}

For the Stop sign, knowing that the sign is red is insufficient to uniquely label the sign given the set of possible labels. Both Yield and Do Not Enter signs are also red. However, knowing the shape of the sign is sufficient to identify the sign as a Stop. We see a similar result for the Hospital sign. The shape of the sign isn't sufficient because a No Pedestrian sign is also a square, but the color, blue, uniquely identifies the sign based on the limited label set.

\begin{figure}[t]
  \center
  \includegraphics[width=0.25\textwidth]{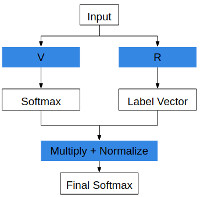}
  \caption{Augmenting an existing classifier $V$ with the robust classifier. The softmax output of $V$ is multiplied with the vector of possible class labels generated by the robust classifier $R$. The result is re-normalized to obtain a new softmax output.}
  \label{fig:augment}
\end{figure}

For the Left Turn Ahead sign, the output vector contains multiple non-zero indices because the extracted features belong to two possible signs. Left Turn Ahead and Right Turn Ahead are both ``yellow'' and ``diamond''. Therefore, the extracted features indicate the output is either of these two sign labels, but cannot specify further. Although this may seem like a limitation, consider the implementation shown in Figure \ref{fig:augment}. The label vector would only allow for adversarial attacks that between classes that share the same set of extracted features. In this example, it is not possible to cause the classifier to label a Stop sign as a Left turn Ahead sign and vice versa.

\ignore{
\subsection{Improving an Existing Classifier}
In general, classification task often expect a single output label for the input. Suppose we have a classifier $V$ that performs well, but may be vulnerable to adversarial examples. Using the basic implementation, we can augment this classifier $V$ and improve its resilience to adversarial examples as shown in Figure \ref{fig:augment}. Namely, adversarial attacks are only possible if find an adversarial example with the same set of extracted features as the original input. In the non-adversarial setting, the classifier would not be negatively affected by the use of resilient feature engineering assuming $R$ is perfectly accurate. Otherwise, the accuracy of the augmented classifier is $acc_V * acc_R$. In the adversarial setting, the attacker can only create adversarial examples with the same set of extracted features as the original input.
}

\section{Conclusion and Future Work}
In this paper, we introduce resilient feature engineering as a methodology for designing adversarially resilient classifiers. Our methodology is supported by the Serial Composition and Parallel Composition theorems, which allow for the chaining of resilient functions to create an adversarially resilient classifier. We also discussed a basic implementation of an adversarially resilient classifier.

However, we leave several open problems for future work. The first problem is identifying predictive features to extract. In this paper, we discussed using pre-existing domain knowledge to identity features useful for classification, but such easily identifiable features are often not present. In other machine learning applications, such as medical imaging or speech recognition, it may be unclear what features are useful. However, work on improving the explainability of learned features as well as determining their relevance to the final prediction may be useful towards feature identification.

A second problem is the feasibility of designing adversarially resilient classifiers. For simpler features, such as color, it may be sufficient to compute the most common color in the input. However, more complex feature such as shape may require more powerful algorithms such as deep neural networks, due to timing constraints or feature complexity. For this, we turn to the existing work in provable adversarial security. Although the existing research focuses providing resilience guarantees for a general classifier, subdividing the classifier into a set of smaller feature extractors may result in more efficient or stronger resilience guarantees. 


\section*{Acknowledgements}
We Haizhong Zheng, Earlence Fernandes, and Somesh Jha for their useful feedback on the ideas in this paper. This work was supported in part by the NSF grant 1646392 and by Didi Labs. Any opinions, findings, and conclusions or recommendations expressed in this material are those of the authors and do not necessarily reflect the views of the National Science Foundation.

{\footnotesize \bibliographystyle{acm}
\bibliography{reference}}

\clearpage

\appendix
\section{Formal Proofs}

\ignore{
\subsection{Proof of Theorem \ref{thm:papr}}
\begin{flalign*}
    & \mbox{$F$ is perfectly accurate} && \mbox{Given} 
    &\\
    &\forall x \in N_F, F(x) = O_F(x) && \mbox{Def of perfectly accurate} 
    &\\
    &N_F = C_F && \mbox{Def of $C_F$ (Section \ref{sec:Classifiers})} 
    &\\
    &N_F = C_F \cup I_F && \mbox{Def of $N_F$ (Section \ref{sec:Classifiers})} &\\
    &N_F/C_F  = (C_F \cup I_F)/C_F && \mbox{Set difference}&\\
    &= \emptyset = C_F/C_F \cup I_F/C_F && \mbox{Set difference is distributive} &\\
    &= I_F = \emptyset && \mbox{$I_F \cap C_F = \emptyset$} \\
    & \forall \lambda > 0, x \in A^\lambda_F \implies x \in I_F && \mbox{Def of $A^\lambda_F$} 
    &\\
    & \forall \lambda > 0, A^\lambda_F \subseteq I_F && \mbox{Def of a subset} 
    &\\
    & \forall \lambda > 0, A^\lambda_F \subseteq \emptyset && \mbox{Substitution} 
    &\\
    & \forall \lambda > 0, A^\lambda_F = \emptyset && \mbox{Subset of  the empty set} 
    &\\
    &  && \mbox{is the empty set} 
    &\\
    & \mbox{F is perfectly resilient} && \mbox{Def of perfectly resilient} 
    &\\
\end{flalign*}
}

\subsection{Proof of Theorem \ref{thm:2class}}
Suppose we have two label sets denoted as $Y$ and $Z$, such that $Y \cap Z = \emptyset$. That is to say, they do not share any labels. Despite this, let us assume it is possible to map some or all of the labels in $Z$ to labels in $Y$. We will define this mapping function as the oracle $O_L$  of the classifier $L$ where $L(z) = y$.
We define two classifiers $F$ and $G$ as such:
\begin{itemize}
    \item $F(x) = y \in Y$
    \item $G(y) = z \in Z$
\end{itemize}
where $F$ is $\lambda$-resilient. Using $F$ and $G$, we also use a classifier $R$ to denote the composition $G(F(x))$ :
\begin{itemize}
    \item $R(x) = G(F(x)) = z \in Z$
    \item $O_R(x) = O_G(O_F(x)) = l \in Z$
\end{itemize}
Note that the mapping function $O_L$ is the inverse function for $O_G$ since both are oracle functions and thus always produce the correct result.

Using these definitions, we restate the Serial Composition Resilience theorem:

\begin{theorem*}
 For arbitrary $z_1,z_2 \in Z$, if $O_L(z_1) = y_1$, $O_L(z_2) = y_2$, $y_1 \neq y_2$, and $R(x) = z_1$, then there is no $\alpha < \lambda$ and $|\gamma| < \alpha$ such that $R(x+\gamma) = z_2$ and $O_R(x+\gamma) = O_R(x) = z_2$
\end{theorem*}

We will verify this theorem using a proof by contradiction and show that  assuming  the opposite of the implication will result in a contradiction with the assumption that F is perfectly resilient.

\begin{flalign*}
    & \exists \alpha < \lambda, 0 < |\gamma| < \alpha:  && \mbox{Assume}
    &\\
    &R(x+\gamma) = z_2 \mbox{ and}  && 
    &\\
    & O_R(x) = O_R(x+\gamma) = z_2  && 
    &\\
    & G(F(x+\gamma)) = z_2 && \mbox{Def of $R$} 
    &\\
    & R(x) = z_1 && \mbox{Given} 
    &\\
    & G(F(x)) = z_1 && \mbox{Def of $R$}
    &\\
    & z_1 \neq z_2 && \mbox{$O_L$ is a function}
    &\\
    & G(F(x)) \neq G(F(x+\gamma)) && \mbox{Substitution}
    &\\
    & F(x) \neq F(x+\gamma) && \mbox{$G$ is a function}
    &\\ 
\end{flalign*}

With this, we have show that the initial assumption has resulted in $F(x) \neq F(x+\gamma)$. To complete the proof, we need to demonstrate that $F(x+\gamma) = O_F(x+\gamma)$. In doing so, it would mean that $x$ is an adversarial example in $A^\alpha_F$ and since $\alpha < \lambda$, $F$ would not be $\lambda$-resilient by definition.

\begin{flalign*}
    & \exists \alpha < \lambda, 0 < |\gamma| < \alpha:  && \mbox{Assume}
    &\\
    &R(x+\gamma) = z_2 \mbox{ and}  && 
    &\\
    & O_R(x) = O_R(x+\gamma) = z_2  && 
    &\\
    & O_L(z_2) = y_2  && \mbox{Given} 
    &\\
    & O_G(O_F(x)) = z_2  && \mbox{Def of $O_R$} 
    &\\
    & O_L(O_G(O_F(x))) = O_L(z_2)  &&
    &\\
    &=O_F(x) = y_2  && \mbox{$O_L$ is the inverse of $O_G$} 
    &\\
    & O_G(O_F(x+\gamma)) = z_2 && \mbox{Def of $O_R$}
    &\\
    & O_L(O_G(O_F(x+\gamma))) = O_L(z_2)  &&
    &\\
    &=O_F(x+\gamma) = y_2    && \mbox{$O_L$ is the inverse of $O_G$} 
    &\\
\end{flalign*}

At this point, we've show that the true label for $x$ and $x+\gamma$ based on the label set $Y$ is $y_2$. What remains for the proof is to establish that $F$ outputs a label for $x$ that is incorrect.
    
\begin{flalign*}
    &R(x+\gamma) = z_2 && \mbox{Given}
    &\\
    &G(F(x+\gamma)) = z_2 && \mbox{Def of $R$}
    &\\
    &G(F(x+\gamma)) = O_G(O_F(x+\gamma)) && \mbox{Substitution}
    &\\
    &F(x+\gamma) = O_F(x+\gamma) && \mbox{Apply $O_L$}
    &\\
    & F(x) \neq F(x+\gamma)  && \mbox{Established previously} 
    &\\
    & F(x) \neq O_F(x)  && \mbox{Substitution} 
    &\\
    & x \in I_F  && \mbox{Def of $I_F$ } 
    &\\
    & x \in A^\lambda_F && \mbox{Def of $A^\alpha_F$} 
    &\\
    & \mbox{$F$ is not $\lambda$-resilient} && \mbox{Def of $\lambda$-resilient}
    &\\
\end{flalign*}

However, it was given that $F$ is perfectly resilient. Therefore, Theorem \ref{thm:2class} is true through proof by contradiction.

\subsection{Proof of Theorem \ref{thm:nclass}}
Suppose we have $n+1$ label sets denoted $Y_1, Y_2$...,$Y_n, Z$ such that $Y_1 \cap Y_2$...$ \cap Y_n \cap Z = \emptyset$. We define the mapping function as the oracle $O_L$ of the classifier $L$ where $L(z) = <y_1, y_2...,y_n>$. $y_i$ denotes a label in the label set $Y_i$.

We define have $n+1$ classifiers such that:
\begin{itemize}
    \item $F_1(x) = y_1 \in Y_1...,F_n(x) = y_n \in Y_n$
    \item $G(<y_1, y_2...,y_n>) = z \in Z$
\end{itemize}
where each $F_i$ for $i=1...n$ is $\lambda$-resilient. Note that $O_L$ is the inverse function of $O_G$. 

Using these classifiers, we use $R$ to denote the composition of $G$ and each $F_i$:
\begin{itemize}
    \item $R(x) = G(F_1(x), F_2(x)..., F_n(x)) = z \in Z$
    \item $O_R(x) = O_G(O_{F_1}(x), O_{F_2}(x)..., O_{F_n}(x)) = l \in Z$
\end{itemize}

Using these definitions, we restate the Parallel Composition resilience theorem:
\begin{theorem*}
 For arbitrary $z_1, z_2 \in Z$, if $O_L(z_1) = C_1$, $O_L(z_2) = C_2$, $C_1 \neq C_2$, and $R(x) = z_1$, then there is no $\alpha < \lambda$ and $|\gamma| < \alpha$ such that $R(x+\gamma) = z_2$ and $O_R(x+\gamma) = O_R(x) = z_2$
\end{theorem*}

For clarification, $C_1, C_2$ are n-tuples and elements of the set $Y_1 x Y_2...x Y_n$. We will verify this theorem using a proof by contradiction and show that assuming the opposite of the implication will result in a contradiction with the assumption that every classifier $F$ is perfectly resilient.

\begin{flalign*}
    & \exists \alpha < \lambda, 0 < |\gamma| < \alpha:  && \mbox{Assume}
    &\\
    & R(x+\gamma) = z_2 \mbox{ and}  && 
    &\\
    & O_R(x) = O_R(x+\gamma) = z_2  && 
    &\\
    & \mbox{Let } S_2 = <F_1(x+\gamma)...,F_n(x+\gamma)> && \mbox{Let}
    &\\ 
    & G(S_2) = z_2 && \mbox{Def of $R$} 
    &\\
    & R(x) = z_1 && \mbox{Given} 
    &\\
    & \mbox{Let } S_1 = <F_1(x)...,F_n(x)> && \mbox{Let}
    &\\ 
    & G(S_1) = z_1 && \mbox{Def of $R$}
    &\\
    & z_1 \neq z_2 && \mbox{Given}
    &\\
    & G(S_1) \neq G(S_2) && \mbox{Substitution}
    &\\
    & S_1 \neq S_2 && \mbox{G is a function}
    &\\
    & \exists i \mbox{ for } 1 \leq i \leq n: && \mbox{Equality of n-tuples}
    &\\
    & F_i(x) \neq F_i(x+\gamma) && \mbox{}
    &\\
\end{flalign*}

With this, we have show that the initial assumption has resulted in $F_i(x) \neq F_i(x+\gamma)$ for some $i$ between 1 and n. We now move to show that $x \in I_{F_i}$ and $x + \gamma \in C_{F_i}$ for that $i$.

\begin{flalign*}
    & \exists \alpha < \lambda, 0 < |\gamma| < \alpha:  && \mbox{Assume}
    &\\
    & R(x+\gamma) = z_2 \mbox{ and}  &&  
    &\\
    & O_R(x) = O_R(x+\gamma) = z_2  && 
    &\\
    & O_L(z_2) = C_2  && \mbox{Given} 
    &\\
    & \mbox{Let } T_1 = <O_{F_1}(x)...,O_{F_n}(x)> && \mbox{Let}
    &\\ 
    & O_G(T_1) = z_2  && \mbox{Def of $O_R$} 
    &\\
    & O_L(O_G(T_1)) = O_L(z_2)  &&
    &\\
    & =T_1 = C_2  && \mbox{$O_L$ is the inverse of $O_G$} 
    &\\
    & \mbox{Let } T_2 = <O_{F_1}(x+\gamma)...,O_{F_n}(x+\gamma)> && \mbox{Let}
    &\\ 
    & O_G(T_2)) = z_2 && \mbox{Def of $O_R$}
    &\\
    & O_L(O_G(T_2)) = O_L(z_2)  &&
    &\\
    & =T_2 = C_2  && \mbox{$O_L$ is the inverse of $O_G$} 
    &\\
    & \mbox{Let } c^i_z \mbox{ denote the ith element of } C_z && \mbox{Let}
    &\\
    & O_{F_i}(x) = O_{F_i}(x+\gamma) = c^i_2    && \mbox{Equality of n-tuples} 
    &\\
\end{flalign*}

Remember that $i$ is the index for the classifier function that is not $\lambda$-resilient. At this point, we've show that the true label for $x$ and $x+\gamma$ based on the label set $Y_i$ is $c^i_2$. What remains for the proof is to establish that $F_i$ outputs a label for $x$ that is incorrect.

\begin{flalign*}
    & R(x+\gamma) = z_2 && \mbox{Given}
    &\\
    & G(S_2) = z_2 && \mbox{Def of $R$ and $S_2$}
    &\\
    & G(S_2) = O_G(T_2) && \mbox{Substitution}
    &\\   
    & S_2 = T_2 && \mbox{Apply $O_L$}
    &\\
    & F_i(x+\gamma) = O_{F_i}(x+\gamma) && \mbox{Apply $O_L$}
    &\\
    & F_i(x) \neq F_i(x+\gamma)  && \mbox{Established previously} 
    &\\
    & F_i(x) \neq O_{F_i}(x)  && \mbox{Substitution} 
    &\\
    & x \in I_{F_i}  && \mbox{Def of $I_{F_i}$ } 
    &\\
    & x \in A^\alpha_{F_i} && \mbox{Def of $A^\alpha_{F_i}$} 
    &\\
    & \mbox{$F_i$ is not $\lambda$-resilient} && \mbox{Def of $\lambda$-resilient} 
    &\\
\end{flalign*}

However, it was given that for $i=1...n, F_i$ is $\lambda$-resilient. Therefore, Theorem \ref{thm:nclass} is true through proof by contradiction.

\end{document}